\documentclass[11pt]{article}
\usepackage{coling2020}
\usepackage{times}
\usepackage{url}
\usepackage{latexsym}
\usepackage{bbding}
\usepackage{enumerate}
\usepackage{booktabs}
\usepackage{multirow}
\usepackage{wasysym}
\usepackage{amssymb}
\usepackage{bm}
\usepackage{graphicx}
\usepackage{makecell}
\usepackage{fancyhdr}

\title{Multiple Generative Models Ensemble for Knowledge-Driven Proactive Human-Computer Dialogue Agent}

\author{Zelin Dai \\
  {\tt circlepi.top@gmail.com} \\
  {\tt Zhejiang Dahua}\\
  {\tt Technology Co., Ltd}\\
  \And
  Weitang Liu \\
  {\tt liuweitangmath@163.com} \\
  {\tt Zhejiang Dahua}\\
  {\tt Technology Co., Ltd}\\
  \And
  Guanhua Zhan \\
  {\tt zhanguanhua@hotmail.com}\\
  {\tt Hangzhou Dianzi}\\
  {\tt University}
  }
\date{}

\begin{document}


\maketitle
\begin{abstract}

  Multiple sequence to sequence models were used to establish an end-to-end multi-turns proactive dialogue generation agent, with the aid of data augmentation techniques and variant encoder-decoder structure designs.  A rank-based ensemble approach was developed for boosting performance.  Results indicate that our single model, in average, makes an obvious improvement in the terms of F1-score and BLEU over the baseline by 18.67\% on the DuConv dataset.  In particular, the ensemble method further significantly outperforms the baseline by 35.85\%.
\end{abstract}

\section{Introduction}
\label{intro}

%
%
    %
    %
    %
    %
    %
    %

Building human-like conversational agent has been recognized as one of the most attractive goals in natural language process (NLP) field, owing to its many promising applications in real life, such as recommendation, custom service and psychological therapy. Benefit from neural network models, an end-to-end and fully data-driven fashion has been more prevalent and advanced in training a conversational agent over the conventional handcrafted rule based method \cite{liu2016not}. However, most existing dialogue systems are still in their early stage, they usually response to queries from users in a completely passive manner rather than a human-like initiative way. In addition, they also have no access to any external knowledge except the conversation history, which makes it challenging for such dialogue systems to produce contentful and factual responses. An active and knowledge-grounded agent would be more desired from the point view both of application and practice, but unfortunately, such dialogue system has been less explored. 

Recent work has shown that retrieval-based and generation-based approaches are two good candidates for building conversation systems. The retrieval-based method ranks all possible responses prepared in advance and choices the best one for given received query, namely query-answer matching. It has the advantage of reasonability, fluency and controllability, but it suffers from the disadvantage of long tail effect and difficulty in customization. Moreover, in practice, it’s usually time-consuming and necessary for manual intervening to prepare the candidate responses set.  The generation-based method overcomes the lacks of the retrieval-based one. However, as pointed by \cite{song2018ensemble}, it usually tends to produce universal and non-informative response, such as ``Yes'', ``I don’t know''. In light of these pros and cons, we proposed an alternative method called generation-reranking. First, the generative models produce some candidate replies. Next, the reranking model is responsible for performing a query-answer matching, to choice a reply as informative as possible over the produced candidates. This method avoids manually constructing the candidate query-answer pairs that will be accomplished by the generative models automatically. The role of the reranking model is similar to the conventional retrieval model, which is, in essence, an ensemble of the generative results.

In this work, we investigated a novel, proactive and knowledge-driven dialogue agent with the generation-reranking method, which has the ability of (1) utilizing the external knowledge to produce response, (2) launching a dialogue even without history turns from human, and (3) leading the dialogue content along given topic goals. As a practical example, we established such a dialogue agent based on the DuConv dataset \cite{wu2019proactive}. It aims to produce proper replies on the condition of given topic goals, related knowledge and history conversation. Our method is fully data-driven, thanks to the encoder-decoder framework. With the aid of text augment technique and multi-runs re-ranking ensemble method, we obtain a notable improvement in performance, by a large margin, over the baseline solution.

\section{Related Works}

Generative method for building conversation chatbots has attracted increasing interest due to its great flexibility. A typical and prevalent generative method is the encoder-decoder \cite{sutskever2014sequence}network, which transforms the input seq{}uence into sequence (Seq2Seq) as output. Iulian et al. used a hierarchical recurrent encoder-decoder neural network to model conversations. Song et al. \cite{song2018ensemble} used multi-Seq2Seq models as generation module of a human-computer conversation system. Both of these systems are designed without external knowledge and in a passive way. Recently, the studies on the dialogue system have been evolved from traditional conversation history based to the knowledge based ones. Marjan et al. \cite{ghazvininejad2018knowledge} generalized the Seq2Seq approach for a knowledge-ground neural conversation model, by conditioning responses on both conversation history and external knowledge. In particular, they employed a facts encoder to inject the external knowledge into the model, similar to the designs by Baidu NLP group \cite{wu2019proactive}. These studies have shown that, the knowledge based system, in general, can produce more meaningful and informative response.

\section{Approaches}

\subsection{Multiple Seq2Seq Network}

Our models follow the Encoder-Decoder framework which has been successfully applied in building a machine translation system and conversation system. In expectation, the models should be capable to produce proper replies based on history conversation and related knowledge, along with the topic goals. Formally, given a input sequence $(x_0,x_1,x_2,...,x_N)$, where the $x_N$ is in the form of (goal, knowledge, history conversation), and the corresponding response sequence $(y_0,y_1,y_2,...,y_t )$ as outputs, the model is aimed at learning the conditional probability distribution $P(y|x,\theta)$, which can be written as:
\begin{equation}
P(y|x,\theta)=\prod_{i=1}^tP(y_i|y_{<i},x_0,...,x_N,\theta)
\end{equation}
where the $\theta$ represents learnable parameters. For ensemble purpose we choose different encoders and decoders for model diversity, i.e., the $\theta$ is parameterized as bidirectional or unidirectional LSTM cells and the Transformer respectively. Besides, we also used some text augmentation techniques for data diversity.

\subsubsection{Data Augmentation}
Data augmentation technique has been widely used in many NLP and computer vision (CV) tasks for improving performance \cite{liu2018feature,wei2019eda}. It’s believed that more data usually enables model to learn more common laws behind a given dataset and thus leads to better generalization ability and performance. In our case, we used the following four text augmentation techniques, and some combination of them to construct multiple datasets.

\begin{itemize}
\item \textbf{Entity Generalization}

Entities, such as person names and industry-specific terminologies, appear frequently in knowledge-driven dialogues.  These discrete entities, even in the same category, have very various forms, resulting in data sparseness that is believed to be a potential reason for overfitting. Our models are supposed to learn the abstract representations of the entity in the specific context, rather than remember the entity itself. Therefore, we generalized the entities of same category with a unique string. For example, we replaced person names ``Ma Long'' with the string of ``person''.
\item \textbf{Knowledge Selection}

We filtered some knowledge which is irrelevant with the current response. Intuitively, this is helpful to reduce the interference from the invalid data. Note that it was done only at the training set for augmentation purpose, because the response is not available at test time anymore.

\item \textbf{Swap}

We randomly swapped the relative position between knowledge and goal, as well as the knowledge (i.e. Subject-Predication-Objects (SPOs) themselves, while remaining the position of conversation history unchanged. 

\item \textbf{Conversation Extraction}

In our case, a single training sample includes multi-turns dialogues. We extracted some child dialogues as training samples with different dialogue turns:
\begin{enumerate}[(1)]
\item all the history dialogue turns
\item the two or three dialogue turns
\end{enumerate}
\end{itemize}

\subsubsection{Model Diversity}
The baseline model \cite{wu2019proactive} is the generation-based network released by Baidu NLP group, which is comprised by the bidirectional gated recurrent unit (GRU) encoder \cite{chung2014empirical}, knowledge manager and the unidirectional decoder. This model is only used for comparison purpose. As stated before, our model is the typical encoder-decoder generator. We fed the goal, conversation history and knowledge together in some organized forms into the encoder. One example of input is schemed in Fig.~\ref{en_de}, where the goal, knowledge and conversation history are spliced by some special flag words that enable the model to distinguish the different data type. We used ``[Goal=N]'' for goal, ``[KG]'' for knowledge, ``[Q=N]'' and ``[A=N]'' for the human turns and machine turns in the dialogues respectively, where N denotes the sequence order. To build a proactive dialogue system, we assumed the start of a conversation would be always ``[Q=0] [CLS]'', a fake utterance from human. By using an encoder, we can get rich representations for the input strings, which could be understood by a decoder to generate a response.
\begin{itemize}
  \item \textbf{LSTM Model}

  Structurally, the LSTM model is similar to that of Nallapati et al. \cite{nallapati2016abstractive} except the layers setting. It contains three parts: the encoder, the decoder and the attention correlation between them. At the encoder side, the tokens are embedded and fed into the LSTM \cite{greff2016lstm} module, outputs as the encoder hidden states $h_i$.  At decoding time, on each timestep $t$, the LSTM decoder receives the embedding of the previous words, and outputs the decoder state $s_{t-1}$. The attention attribution is calculated over the $h_i$ and $s_{t-1}$ to tell the decoder which words at the encoder end should be focused on, and then produces the vocabulary distribution $P_{vocab}$ over all words in the vocabulary:
   \begin{equation}
  P_{vocab}=softmax(W'(W[s_{t-1},h_t^{*}+b]+b'))
  \end{equation}
  where $W^{'}$, $W$, $b$ and $b'$ are learnable parameters.  The $h_t^*$ is the well-known context vector, which can be expressed as:
  \begin{equation}
  h_t^{*} = \sum_ia_i^{t}h_i
  \end{equation}
  where the $a_i^t$ denotes the similarity between $h_i$ and $s_{t-1}$.

  Practically, we choose several combinations of the number of LSTM layers for the encoder and decoder to increase model diversity. Besides, in some cases the models were trained with a pre-trained word embedding, while in the other cases, the embedding layer was randomly initialized. Our model structures were summarized in Table~\ref{tab1}.  Detailed descriptions about the hyper-parameter settings at training and test time will be explained in the experiment section later.
\begin{figure}[htbp] 
\centering 
\includegraphics[height=6.5cm, width=15cm]{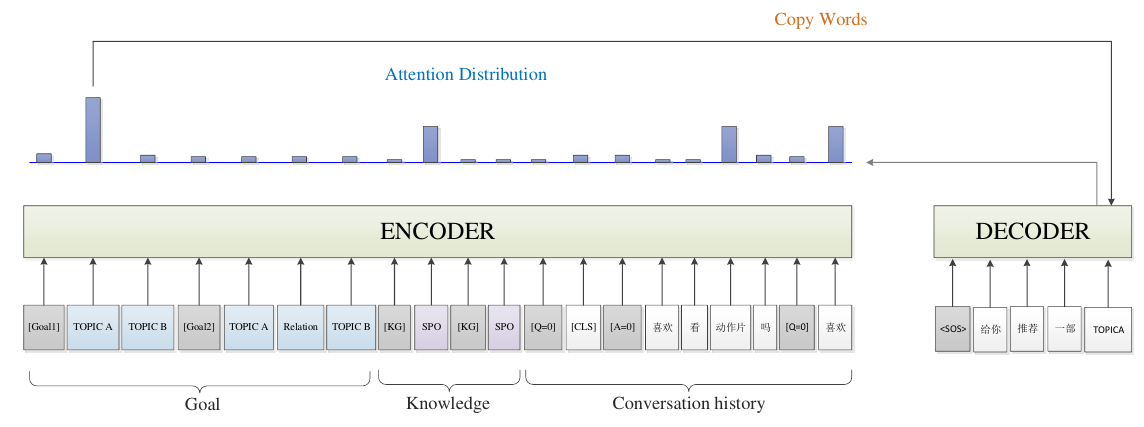} 
\caption{The encoder-decoder model structure used in this work} 
\label{en_de}
\end{figure}
  \item \textbf{Transformer Model}

  Our transformer model \cite{vaswani2017attention} follows the ones typically used in machine translation system, which is based on multi-layer self-attention and has achieved state-of-the-art results in many sequence to sequence tasks.
  \begin{table}[h]
   \caption{\label{font-table} multiple model structures used in this work }
  \begin{center}
  \begin{tabular}{c|c|c|c}
  \hline \bf Model & \bf Encoder & \bf Decoder & \bf Decoder \\ \hline
  LSTM-L11 & 1 layer Bidirectional LSTM & 1 layer unidirectional LSTM & YES/NO \\
  LSTM-L31 & 3 layers Bidirectional LSTM & 1 layer unidirectional LSTM & NO \\
  LSTM-L22 & 2 layers Bidirectional LSTM & 2 layers unidirectional LSTM & NO\\
  Transformer & Transformer (6 layers) & Transformer (6 layers) & NO \\
  \hline
  \end{tabular}
  \label{tab1}
  \end{center}
  \end{table}
 
\end{itemize}

\subsubsection{Copy Mechanism}
\begin{figure}[htbp] 
\centering 
\includegraphics[height=8cm, width=15cm]{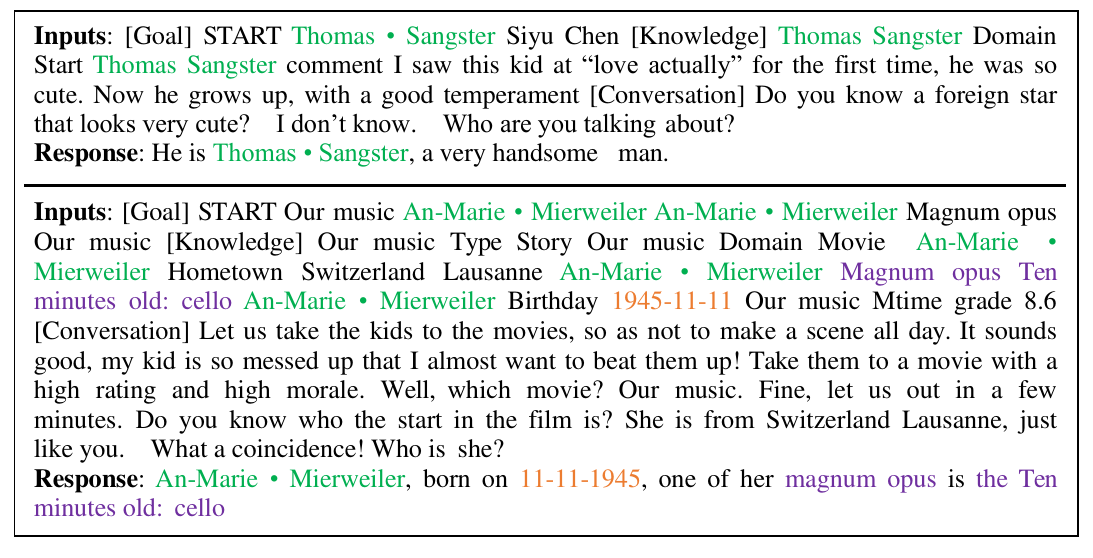} 
\caption{(Translated) Examples for illustrating the knowledge reuse in the response.  The colored words in the response are directly copying from the knowledge.} 
\label{sentence}
\end{figure}

For a knowledge-driven dialogue chatbot, an important ability is to reuse the given knowledge to produce replies.  As illustrated in Fig.~\ref{sentence}, the response references the knowledge in the way of copying some words of it.  To accomplish this, a pointer-generator network \cite{see2017get} was employed at the decoder end. The pointer-generator allows the model to selectively copy words from knowledge while retaining the ability to generate a new word.  Specifically, it uses a generation probability $P_{gen}$ as a controller to determine the model when to extract word from knowledge and when to generate a new word from the vocabulary. The $P_{gen}$ is formally defined as:
\begin{equation}
P_{gen} = sigmoid(W_hh_t^*+W_ss_{t-1}+W_xx_t+b)
\end{equation}
where the $W_h$, $W_s$, $W_x$, $b$ are learnable parameters. For each training sample, it preserves a dynamic word dictionary which is comprised by the fix vocabulary and the words of present inputs at encoder end. The probability distribution over the dynamic word dictionary, which is different from the equation (2), can be written as:
\begin{equation}
P(w)=P_{gen}P_{vocab}(w)+(1-P_{gen}\sum_{i:w_i=w}a_i^t)
\end{equation}
Similar to the machine summary tasks, the copy mechanism is aimed at trade off between generation and extraction.

\subsection{Rank-based ensemble}

We employed a Gradient Boosting Decision Tree (GBDT) regressor \cite{friedman2001greedy} to score the all output responses by the encoder-decoder models, ranked and selected the best one as the final response.  This is based on the observations that a model could perform well overall, but sometimes gets bad result in certain instances, and vice versa.  It’s expected that the relatively weak models can compensate each to others and form a stronger model.  We extracted some high-level structural features from the output responses, as listed in the following:
\begin{enumerate}[(1)]
\item Model weights. Response from different model was distinguished by a prior model weight, which is determined by the model’s overall performances at the develop set. 
\item Length. Too short or too long response is usually undesired. Therefore we characterized the length of a sentence in the word level and char level, respectively.
\item Similarity. We calculated the N-gram BLEU similarity, semantic similarity (i.e. Word2vec similarity, TF-IDF similarity), and topic similarity (i.e. LSA similarity) between the inputs, especially the closest conversation history and the response.
\item Entity statistics. Entity is a very key element in the knowledge-based replies. We count the entity frequency, possible and impossible entity co-occurrence times in the response that reflects the consistency between the prior knowledge and response.
\item Sentence fluency. We simply defined the sentence fluency by the reciprocal of N-gram repeat counts. A more fluent response is more preferred.
\item Beam score. The score of the best path for generating the response sequence by beam search algorithm.
\end{enumerate}
\section{Experiments}
\subsection{Datasets}
We used the Chinese DuConv conversation dataset, which includes 30k dialogues, about 120k dialogue turns, of which 20k dialogues are training set, 2k dialogues are development set and 5k dialogues are used for testing. Each dialogue is comprised by three parts:
\begin{enumerate}[(1)]
\item Dialogue goal, it contains two or three lines: the first contains the given dialogue path, i.e. [``Start'', ``Topic-A,'' ``Topic-B'']. The other lines contain the relationship of Topic-A and Topic-B.
\item Knowledge, background knowledge related to Topic-A and Topic-B, which is in the form of SPOs.
\item Dialogue history, conversation sequences before the current utterance, it's empty if the current utterance is the start of the conversation.
\end{enumerate}
\subsection{Evaluation Method}
To evaluate the results, we used the F1 (char-based F-score of output responses against gold responses) and BLEU (BLEU1 and BLEU2, word-based precision of output responses against gold responses). The total score is calculated by: score = F1 + BLEU1 + BLEU2. 
\subsection{Experimental Details}
\subsubsection{Data preparation}
By using text augment techniques, we constructed five datasets, as listed in Table~\ref{tab2}, for the training of the generative models.  These datasets are comprised by $<Q,R>$ pairs, where $Q$ denotes the concatenation of goal, knowledge and conversation history, the $R$ denotes the corresponding reply.  As can be seen in Table~\ref{tab2}, some training samples are shared across the datasets while keeping the differences between the datasets.  After preprocessing, the develop set and test set includes 10k and 5k samples, respectively. 
\begin{table*}[htbp]
\caption{The prepared data for the generative models}
\begin{center}
\setlength{\tabcolsep}{2mm}{
\begin{tabular}{c|c|c|c|c}
\hline
\multirow{2}{*}{Data} &\multicolumn{4}{|c}{Text Augment Operation} \\
\cline{2-5}
& Entity Generalization & Knowledge Selection & Switch & Conversation Extraction \\
\cline{2-5} 
\hline
D-1 & \XSolidBrush & \XSolidBrush & \XSolidBrush & \Checkmark(all history turns) \\
\hline
D-2 & \Checkmark & \XSolidBrush & \XSolidBrush & \Checkmark(all history turns)\\
\hline
D-3 & \Checkmark & \Checkmark & \XSolidBrush & \Checkmark(two, three \& all history turns)\\
\hline
D-4 & \Checkmark & \Checkmark & \XSolidBrush & \Checkmark(all history turns)\\
\hline
D-5 & \Checkmark & \XSolidBrush & \Checkmark & \Checkmark(two, three \& all history turns)\\
\hline
D-6 & \Checkmark & \Checkmark & \Checkmark & \Checkmark(all history turns)\\
\hline
\end{tabular}}
\label{tab2}
\end{center}
\end{table*}
\subsubsection{Model Training \& Testing}
With the multiple model structures and datasets together, we trained 27 generative models for ensemble.  All the model training and testing were performed on the NVIDIA 2070 GPU. The LSTM models were trained with a hidden size of 256 and a batch size of 16. In the case of training with pre-trained word embedding, we used an open source Chinese pre-trained word embedding \cite{klein2017opennmt} with the dimension of 200 to initialize the embedding layer.  In the other cases, the dimension of word embedding is set to 128. They are optimized with the Adam optimizer at a learning rate of 0.001 and further fine-tuning at 0.0001. The transformer models were trained with the standard parameters setting as described in \cite{song2018directional}. We applied an early stopping operation during training when the perplexity on the develop set has no tendency to decrease.

At test time, the model decodes using the beam search algorithm. We introduce coverage penalty and length normalization mechanism to the conventional max-probability based beam search algorithm. The beam size, length normalization factor and coverage penalty constant were optimized on the develop set with the metrics of score described in 4.2.
\subsubsection{Model Ensemble}
After decoding, we have all 27 model’s responses for each sample. For training the ranker, we constructed a new supervised dataset based on the develop set. Specifically, we score all candidate responses for each develop sample by calculating the sum of F1 score, BLEU1 and BLEU2 over the ground truth. As a result, the new dataset includes 10k $\times$ 27 samples. The ranker training was performed using the Xgboost package \cite{chen2016xgboost}, with 1900 regression trees and a depth of 6. For the test set, we use the trained ranker to find the one that has the highest score in all the candidates as the final response.
\subsection{Results}
As shown in Table~\ref{tab3}, we list all the single models’ result and ensemble result, as well as the official baseline result on the test set. It’s clear that our single models significantly outperform the baseline model.  The average score of the all single models reaches 98.12\%, 18.67\% higher than the baseline. In particular, the best single model (Transformer D-5) has a score of 102.82\%, which is 23.37\% higher than the baseline. These results indicate that the solution we proposed is proper for building the dialogue agent. The transformer models, in general, have better performance than that of LSTM models.  Besides, it can be seen in Table~\ref{tab3} that the more text augment techniques are used, often the better performance the transformer model has. However, this phenomenon was not observed in the LSTM models. The diverse models, together with the multiple dataset bring a further boosting, the score is 35.85\% higher over the baseline and 12.48\% improvement over the best single model, which is consistent with our expectation.  
\begin{table*}[htbp]
\caption{Our results on the test set. The evaluation metrics (unit :\%) are given in the form of Score [F1+BLEU1+BLEU2].}
\begin{center}
\resizebox{\textwidth}{!}{
\begin{tabular}{c|c|c|c|c|c}
  \hline
  & LSTM-L11 & LSTM-L11-Embed & LSTM-L22 & LSTM-L31 & Transformer\\
  \hline
  D-1 & \makecell*[c]{96.41\\ $[42.22+32.5+21.7]$} 
      & \makecell*[c]{96.67\\ $[42.36+32.5+21.8]$} 
      & \makecell*[c]{97.96\\ $[42.85+32.9+22.2]$} 
      & \makecell*[c]{97.90\\ $[42.45+33.2+22.2]$} 
      & /
      \\
  \hline
  D-2 & \makecell*[c]{98.63\\ $[42.79+33.4+22.5]$} 
      & \makecell*[c]{98.41\\ $[42.88+33.2+22.3]$} 
      & \makecell*[c]{98.51\\ $[42.89+33.3+22.3]$} 
      & \makecell*[c]{97.83\\ $[42.73+33.0+22.1]$} 
      & \makecell*[c]{98.02\\ $[41.52+34.1+22.4]$} 
      \\
  \hline
  D-3 & \makecell*[c]{98.23\\ $[42.71+33.3+22.3]$} 
      & \makecell*[c]{96.81\\ $[42.26+32.7+21.8]$} 
      & \makecell*[c]{95.41\\ $[41.81+32.4+21.2]$} 
      & /
      & \makecell*[c]{100.89\\ $[42.50+35.0+23.4]$} 
      \\
  \hline
  D-4 & \makecell*[c]{97.28\\ $[42.29+33.0+22.0]$} 
      & \makecell*[c]{96.91\\ $[42.38+32.7+21.9]$} 
      & \makecell*[c]{97.40\\ $[42.73+32.7+21.9]$} 
      & \makecell*[c]{97.14\\ $[42.25+32.9+22.0]$} 
      & \makecell*[c]{102.00\\ $[42.21+36.1+23.7]$} 
      \\
  \hline
  D-5 & \makecell*[c]{98.72\\ $[42.15+33.9+22.6]$} 
      & \makecell*[c]{95.88\\ $[41.98+32.3+21.7]$} 
      & \makecell*[c]{97.65\\ $[42.29+33.2+22.1]$} 
      & \makecell*[c]{97.37\\ $[42.26+33.0+22.1]$} 
      & \makecell*[c]{$\bf{102.82}$\\ $[43.18+35.8+23.8]$} 
      \\
  \hline
  D-6 & \makecell*[c]{96.99\\ $[42.07+33.1+21.8]$} 
      & \makecell*[c]{97.72\\ $[42.57+33.0+22.1]$} 
      & \makecell*[c]{98.61\\ $[42.9+33.5+22.3]$} 
      & /
      & \makecell*[c]{101.12\\ $[42.04+35.7+23.4]$} 
      \\
  \hline
  Ensemble &  &  &  \makecell*[c]{$\bf{115.3}$\\ $[46.2+41.5+27.6]$} & & \\
  \hline  
  Official baseline &  &  &  \makecell*[c]{$\bf{79.45}$\\ $[32.65+30.0+16.8]$} & & \\
  \hline 
\end{tabular}}
\end{center}
\label{tab3}
\end{table*}

\begin{figure}[htbp] 
\centering 
\includegraphics[height=8cm, width=15cm]{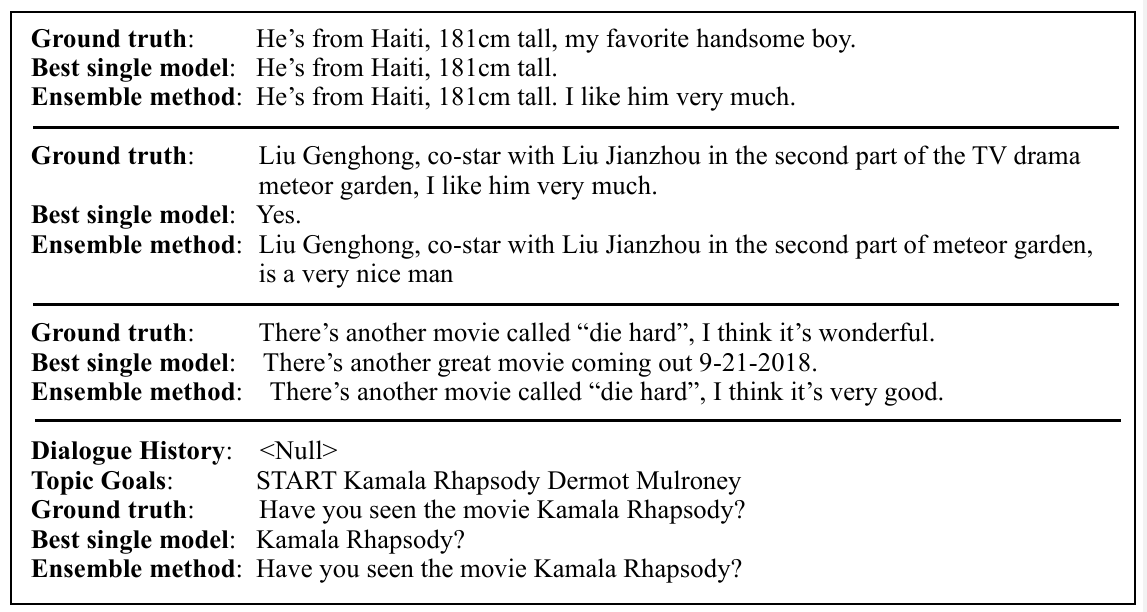} 
\caption{(Translated) A comparison of result between the best single model and the ensemble method. ``$<$Null$>$'' means there’s no any dialogue history.} 
\label{sentence_2}
\end{figure}

Fig.~\ref{sentence_2} illustrates several examples to make a comparison between the ensemble result and the best single model result. It can be seen that, comparing with the best single model, the ensemble method prefers to generate more informative and complete replies, which is attributed to the re-ranking model. While there still some differences in words between the ensemble result and the ground truth, it’s fair to say that they indeed share the similar semantics, which can meet the requirement of a human-computer dialogue agent. It’s worth noting that, even without any dialogue history, the designed agent is able to launching a dialogue along with the given topic goals, as shown the last example in Fig.~\ref{sentence_2}, which is one of distinctive advantage of our dialogue agent.

\section{Conclusion}
In this work, we presented a proactive, knowledge based, fully data driven dialogue agent.  Multiple generative models were designed to produce candidate replies and a re-ranking model was developed for retrieving the best one from the produced candidates. The proposed method obviously outperforms the baseline by a large margin, which provides valuable reference for building an end-to-end highly- performed human-computer dialogue system.

\section*{Acknowledgment}
The authors would like to acknowledge Baidu NLP for providing the open dialogue datasets and the baseline.  


\bibliographystyle{acl}
\bibliography{ref.bib}

\begin{thebibliography}{}

\bibitem[\protect\citename{Chen and Guestrin}2016]{chen2016xgboost}
Tianqi Chen and Carlos Guestrin.
\newblock 2016.
\newblock Xgboost: A scalable tree boosting system.
\newblock In {\em Proceedings of the 22nd acm sigkdd international conference
  on knowledge discovery and data mining}, pages 785--794. ACM.

\bibitem[\protect\citename{Chung \bgroup et al.\egroup
  }2014]{chung2014empirical}
Junyoung Chung, Caglar Gulcehre, KyungHyun Cho, and Yoshua Bengio.
\newblock 2014.
\newblock Empirical evaluation of gated recurrent neural networks on sequence
  modeling.
\newblock {\em arXiv preprint arXiv:1412.3555}.

\bibitem[\protect\citename{Friedman}2001]{friedman2001greedy}
Jerome~H Friedman.
\newblock 2001.
\newblock Greedy function approximation: a gradient boosting machine.
\newblock {\em Annals of statistics}, pages 1189--1232.

\bibitem[\protect\citename{Ghazvininejad \bgroup et al.\egroup
  }2018]{ghazvininejad2018knowledge}
Marjan Ghazvininejad, Chris Brockett, Ming-Wei Chang, Bill Dolan, Jianfeng Gao,
  Wen-tau Yih, and Michel Galley.
\newblock 2018.
\newblock A knowledge-grounded neural conversation model.
\newblock In {\em Thirty-Second AAAI Conference on Artificial Intelligence}.

\bibitem[\protect\citename{Greff \bgroup et al.\egroup }2016]{greff2016lstm}
Klaus Greff, Rupesh~K Srivastava, Jan Koutn{\'\i}k, Bas~R Steunebrink, and
  J{\"u}rgen Schmidhuber.
\newblock 2016.
\newblock Lstm: A search space odyssey.
\newblock {\em IEEE transactions on neural networks and learning systems},
  28(10):2222--2232.

\bibitem[\protect\citename{Klein \bgroup et al.\egroup }2017]{klein2017opennmt}
Guillaume Klein, Yoon Kim, Yuntian Deng, Jean Senellart, and Alexander~M Rush.
\newblock 2017.
\newblock Opennmt: Open-source toolkit for neural machine translation.
\newblock {\em arXiv preprint arXiv:1701.02810}.

\bibitem[\protect\citename{Liu \bgroup et al.\egroup }2016]{liu2016not}
Chia-Wei Liu, Ryan Lowe, Iulian~V Serban, Michael Noseworthy, Laurent Charlin,
  and Joelle Pineau.
\newblock 2016.
\newblock How not to evaluate your dialogue system: An empirical study of
  unsupervised evaluation metrics for dialogue response generation.
\newblock {\em arXiv preprint arXiv:1603.08023}.

\bibitem[\protect\citename{Liu \bgroup et al.\egroup }2018]{liu2018feature}
Bo~Liu, Xudong Wang, Mandar Dixit, Roland Kwitt, and Nuno Vasconcelos.
\newblock 2018.
\newblock Feature space transfer for data augmentation.
\newblock In {\em Proceedings of the IEEE Conference on Computer Vision and
  Pattern Recognition}, pages 9090--9098.

\bibitem[\protect\citename{Nallapati \bgroup et al.\egroup
  }2016]{nallapati2016abstractive}
Ramesh Nallapati, Bowen Zhou, Caglar Gulcehre, Bing Xiang, et~al.
\newblock 2016.
\newblock Abstractive text summarization using sequence-to-sequence rnns and
  beyond.
\newblock {\em arXiv preprint arXiv:1602.06023}.

\bibitem[\protect\citename{See \bgroup et al.\egroup }2017]{see2017get}
Abigail See, Peter~J Liu, and Christopher~D Manning.
\newblock 2017.
\newblock Get to the point: Summarization with pointer-generator networks.
\newblock {\em arXiv preprint arXiv:1704.04368}.

\bibitem[\protect\citename{Song \bgroup et al.\egroup
  }2018a]{song2018directional}
Yan Song, Shuming Shi, Jing Li, and Haisong Zhang.
\newblock 2018a.
\newblock Directional skip-gram: Explicitly distinguishing left and right
  context for word embeddings.
\newblock In {\em Proceedings of the 2018 Conference of the North American
  Chapter of the Association for Computational Linguistics: Human Language
  Technologies, Volume 2 (Short Papers)}, pages 175--180.

\bibitem[\protect\citename{Song \bgroup et al.\egroup }2018b]{song2018ensemble}
Yiping Song, Rui Yan, Cheng-Te Li, Jian-Yun Nie, Ming Zhang, and Dongyan Zhao.
\newblock 2018b.
\newblock An ensemble of retrieval-based and generation-based human-computer
  conversation systems.

\bibitem[\protect\citename{Sutskever \bgroup et al.\egroup
  }2014]{sutskever2014sequence}
Ilya Sutskever, Oriol Vinyals, and Quoc~V Le.
\newblock 2014.
\newblock Sequence to sequence learning with neural networks.
\newblock In {\em Advances in neural information processing systems}, pages
  3104--3112.

\bibitem[\protect\citename{Vaswani \bgroup et al.\egroup
  }2017]{vaswani2017attention}
Ashish Vaswani, Noam Shazeer, Niki Parmar, Jakob Uszkoreit, Llion Jones,
  Aidan~N Gomez, {\L}ukasz Kaiser, and Illia Polosukhin.
\newblock 2017.
\newblock Attention is all you need.
\newblock In {\em Advances in neural information processing systems}, pages
  5998--6008.

\bibitem[\protect\citename{Wei and Zou}2019]{wei2019eda}
Jason~W Wei and Kai Zou.
\newblock 2019.
\newblock Eda: Easy data augmentation techniques for boosting performance on
  text classification tasks.
\newblock {\em arXiv preprint arXiv:1901.11196}.

\bibitem[\protect\citename{Wu \bgroup et al.\egroup }2019]{wu2019proactive}
Wenquan Wu, Zhen Guo, Xiangyang Zhou, Hua Wu, Xiyuan Zhang, Rongzhong Lian, and
  Haifeng Wang.
\newblock 2019.
\newblock Proactive human-machine conversation with explicit conversation
  goals.
\newblock {\em arXiv preprint arXiv:1906.05572}.

\end{thebibliography}

\end{document}